\documentclass[10pt,journal,compsoc]{IEEEtran}
\ifCLASSOPTIONcompsoc
  \usepackage[nocompress]{cite}
\else
  \usepackage{cite}
\fi
\ifCLASSINFOpdf
   \usepackage[pdftex]{graphicx}
   \DeclareGraphicsExtensions{.pdf,.jpeg,.jpg,.png}
\else
   \usepackage[dvips]{graphicx}
   \DeclareGraphicsExtensions{.eps}
\fi
\graphicspath{{figures/}{pictures/}{images/}{./}} %
\usepackage{microtype}                 % use micro-typography (slightly more compact, better to read)
\PassOptionsToPackage{warn}{textcomp}  % to address font issues with \textrightarrow
\usepackage{textcomp}                  % use better special symbols
\usepackage{mathptmx}                  % use matching math font
\usepackage{times}                     % we use Times as the main font
\usepackage{cite}                      % needed to automatically sort the references
\usepackage{tabu}                      % only used for the table example
\usepackage{booktabs}                  % only used for the table example

\usepackage{algorithm}
\usepackage{algorithmic}

\usepackage{paralist}
\usepackage{enumitem}
\usepackage[colorlinks,linkcolor=blue]{hyperref}
\usepackage{subfigure}
\usepackage{wrapfig}
\usepackage{xcolor}

\newcommand{\rev}[1]{{\color{black}#1}}
\usepackage{dblfloatfix}
\begin{document}
\title{VAC-CNN: A Visual Analytics System \\for Comparative Studies of Deep \\Convolutional Neural Networks}
% \title{VAC-CNN: A Visual Analytics System \\for Comparative Studies of Deep CNNs}

\author{Xiwei~Xuan,
        Xiaoyu~Zhang,
        Oh-Hyun~Kwon,
        and~Kwan-Liu~Ma% <-this % stops a space
\IEEEcompsocitemizethanks{\IEEEcompsocthanksitem X. Xuan, X. Zhang, O.-H. Kwon and K.-L. Ma are with University of California, Davis.\protect\\
    E-mail: \{xwxuan, xybzhang, kw, klma\}@ucdavis.edu.}% <-this % stops an unwanted space
\thanks{Manuscript received October 22, 2021; revised January 14, 2022.}}


\markboth{Journal of \LaTeX\ Class Files,~Vol.~14, No.~8, August~2015}%
{Shell \MakeLowercase{\textit{et al.}}: Bare Demo of IEEEtran.cls for Computer Society Journals}



\IEEEtitleabstractindextext{%
\begin{abstract}
% The rapid development of Convolutional Neural Networks (CNNs) in recent years has triggered significant breakthroughs in many machine learning (ML) applications. The ability to understand and compare various CNN models available is thus essential. The conventional approach with visualizing each model's quantitative features, such as classification accuracy and computational complexity, is not sufficient for a deeper understanding and comparison of the behaviors of different models. Moreover, most of the existing tools for assessing CNN behaviors only support comparison between two models and lack the flexibility of customizing the analysis tasks according to user needs. This paper presents a visual analytics system, CNN Comparator (CNNC), that supports the in-depth inspection of a single CNN model as well as comparative studies of two or more models. The ability to compare a larger number of (e.g., tens of) models especially distinguishes our system from previous ones. With a carefully designed model visualization and explaining support, CNNC facilitates a highly interactive workflow that promptly presents both quantitative and qualitative information at each analysis stage. We demonstrate CNNC's effectiveness for assisting ML practitioners in evaluating and comparing multiple CNN models through two use cases and one preliminary evaluation study using the image classification tasks on the ImageNet dataset.
The rapid development of Convolutional Neural Networks (CNNs) in recent years has triggered significant breakthroughs in many machine learning (ML) applications. The ability to understand and compare various CNN models available is thus essential. The conventional approach with visualizing each model's quantitative features, such as classification accuracy and computational complexity, is not sufficient for a deeper understanding and comparison of the behaviors of different models. Moreover, most of the existing tools for assessing CNN behaviors only support comparison between two models and lack the flexibility of customizing the analysis tasks according to user needs. This paper presents a visual analytics system, VAC-CNN (\underline{V}isual \underline{A}nalytics for \underline{C}omparing \underline{C}NNs), that supports the in-depth inspection of a single CNN model as well as comparative studies of two or more models. The ability to compare a larger number of (e.g., tens of) models especially distinguishes our system from previous ones. With a carefully designed model visualization and explaining support, VAC-CNN facilitates a highly interactive workflow that promptly presents both quantitative and qualitative information at each analysis stage. We demonstrate VAC-CNN's effectiveness for assisting novice ML practitioners in evaluating and comparing multiple CNN models through two use cases and one preliminary evaluation study using the image classification tasks on the ImageNet dataset.
\end{abstract}

}

\maketitle

\IEEEdisplaynontitleabstractindextext

% \fi
\IEEEpeerreviewmaketitle

\IEEEraisesectionheading{\section{Introduction}\label{sec:introduction}}
\begin{figure*}[ht]
  \centering
  \includegraphics[width=1\textwidth]{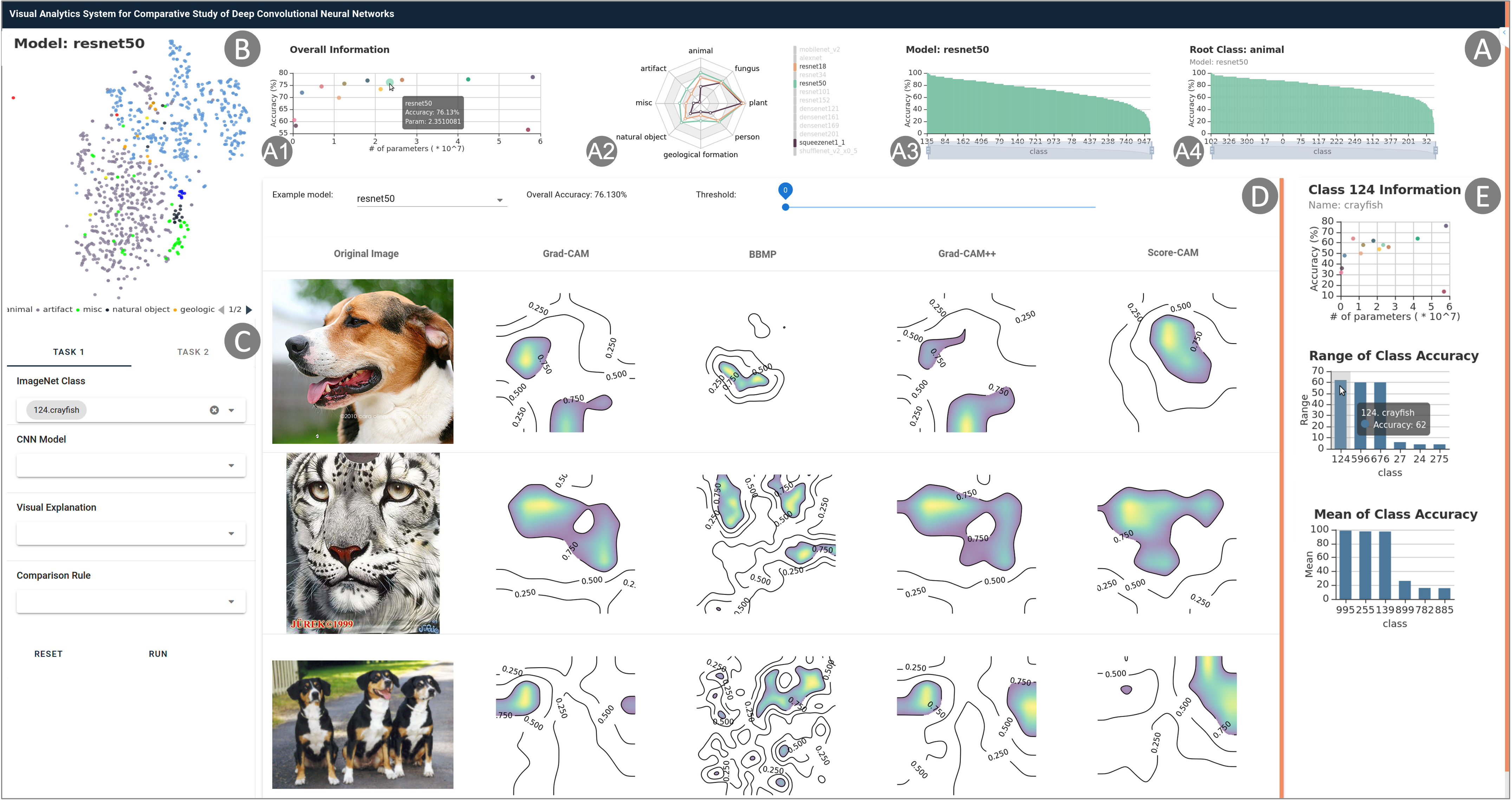}
  \caption{
  The interface of our visual analytics system.
  \textbf{(A) Overall Information View} visualizes multiple quantitative matrices of each CNN model.
  \textbf{(B) Distribution Graph View} shows the aggregating activations of $1000$ ImageNet classes generated from the distance matrix of each model's predictions.
  \textbf{(C) Task Selection Sidebar} facilitates users to customize multiple selections to interactively compare the models' behaviors.
  \textbf{(D) Visual Explanation View} provides two types of information - before running any task, it presents the visual explanation examples; when users are running a task, it shows the \rev{explanation} results.
  \textbf{(E) Supplemental View} works as a supplementary information board.
  }
  \label{fig:interface}
\end{figure*}

% \IEEEPARstart{I}{n} recent years, machine learning (ML) researchers have made great efforts to boost the performance of computer vision tasks by taking advantage of the state-of-the-art deep Convolutional Neural Networks (CNNs) \cite{lecun1995learning,krizhevsky2012imagenet,he2016deep,iandola2016squeezenet,huang2017densely,howard2017mobilenets,zhang2018shufflenet,hu2018competitive}.
\IEEEPARstart{I}{n} recent years, \rev{researchers have pushed the boundary of various domains unprecedentedly} by taking advantage of the state-of-the-art deep Convolutional Neural Networks (CNNs) \cite{bhatt2021cnn,lecun1995learning,krizhevsky2012imagenet,he2016deep,iandola2016squeezenet,huang2017densely,howard2017mobilenets,zhang2018shufflenet,hu2018competitive}.
\rev{During this process, many machine learning (ML) practitioners with diverse knowlege backgrounds share the common need to understand and compare multiple CNNs.}
Such comparison tasks \rev{are challenging for novice ML practitioners who have primary but not comprehensive ML knowledge background}, especially when \rev{the number of models to compare is large and} the features of them vary a lot.
\rev{For example, a medical school graduate student may want to adopt a CNN for disease detection.
With tens of different CNN architectures available, it is difficult for them to filter out inapplicable models, let alone to find one with desired features.}
Conventional approaches for comparing multiple CNNs \cite{aydogdu2017comparison,liu2017survey,khan2019survey,canziani2016analysis,mukhopadhyay2019performance} often focus on investigating model architectures \cite{aydogdu2017comparison,liu2017survey,khan2019survey} or analyzing quantitative performances statically \cite{canziani2016analysis,aydogdu2017comparison,mukhopadhyay2019performance}, but fail to provide enough intuitive information or reasons behind the different behavior of models. 
Therefore, it calls for efforts to develop novice-friendly tools for ML practitioners that improve models' transparency, reveal models' differences, and extend models' applications through understanding their behaviors in CNN comparative studies.

An interpretable CNN comparative study can be divided into two phases---model interpretation and model comparison.
For model interpretation, researchers from the XAI (eXplainable Artificial Intelligence) \cite{gunning2017explainable} community have developed plenty of class-discriminative visual explanation methods as a post-hoc analysis of the underlying behaviors of deep models \cite{zhou2016learning,selvaraju2017grad,fong2017interpretable,chattopadhay2018grad,omeiza2019smooth,wang2019score}.
These methods highlight the region of interest (ROI) relevant to the model's decision, and could significantly increase the interpretability of deep models \cite{zhou2016learning}.
However, most of them are only applied to analyze a single model's behaviors in detail, while rarely used to compare multiple models. 
For model comparison, many visual analytics tools have been developed for interactive CNN comparison \cite{krause2016interacting, zeng2017cnncomparator, zhang2018manifold, das2019beames, murugesan2019deepcompare, arendt2020parallel}.
They integrate different visualization techniques to compare deep models from different perspectives, such as feature activations, parameter distributions, etc.
Some of these tools support multi-model comparison, but they either lack interpretability \cite{krause2016interacting, zeng2017cnncomparator, zhang2018manifold, das2019beames} or only support comparison between two models \cite{zeng2017cnncomparator, murugesan2019deepcompare, arendt2020parallel}.
In response to the increasing number of models to compare and choose from, it is necessary to consolidate the state-of-the-art techniques from both phases and develop a CNN model comparative study tool that can take a flexible number of models and provide explanations for model behavior.

In this paper, we introduce a visual analytics system---VAC-CNN (\underline{V}isual \underline{A}nalytics for \underline{C}omparing \underline{C}NNs)---to support an interpretable comparative study of deep CNNs.
VAC-CNN assists the progress of a highly interactive workflow with carefully designed visualizations.
To facilitate flexible comparison customization, VAC-CNN supports three types of comparison studies: 1) high-level screening for a large number of (e.g. tens of) models, 2) behavior consistencies evaluation for a few models, and 3) detailed investigation for a single model.
To enhance models' interpretability, VAC-CNN integrates multiple class-discriminative visual explanation methods, including Grad-CAM\cite{selvaraju2017grad}, BBMP\cite{fong2017interpretable}, Grad-CAM++\cite{chattopadhay2018grad}, Smooth Grad-CAM++\cite{omeiza2019smooth}, and Score-CAM\cite{wang2019score}.
To present the results of these methods smoothly, VAC-CNN promptly visualizes both quantitative and qualitative information at each analysis stage, allowing users to investigate and compare multiple models from different perspectives.

We illustrate the effectiveness of our visualization and interaction design to assist ML novices in CNN interpretation and comparison with two use cases.
One is about multi-model comparison on a single input image, and the other is about single model behavior inspection on different classes of images.
We also evaluate the usefulness of VAC-CNN with a preliminary evaluation study.
According to the evaluation result, our system is easy to use and capable of providing useful insights about model behavior patterns for novice ML practitioners.

The primary contributions of our work include:
\begin{itemize}[topsep=0pt,parsep=0pt,itemsep=0pt,partopsep=0pt,itemindent=0pt,leftmargin=10pt]
\item A visual analytics system to support flexible CNN model analysis from single-model inspection to multi-model comparative study.
\item A suite of enhanced visual explanation methods coordinated by a highly interactive workflow for effective and interpretable model comparison.
\end{itemize}

\section{Related Work}
\label{sec:RelatedWork}
Our system for the comparative study of interpretable CNN models is inspired by previous works related to deep learning and XAI.
This section discusses existing research on visual explanation methods for understanding CNN model behaviors, CNN model comparison, and visual analytics for interpretable CNN comparison.

% Visual Explanation Methods
\subsection{Visual Explanation Methods for Interpretable CNN}
Visual explanation methods play an essential role in improving the transparency of deep CNN models.
According to the visualization purpose, the existing visual explanation methods can be grouped into three kinds.
The first group of methods mainly focus on visualizing the activations of neurons and layers inside a specific model, such as Feature Visualization\cite{olah2017feature} and Deep Dream\cite{mordvintsev2015deepdream}.
These methods focus on exploring a single model's internal operation mechanism, which is not scalable for comparing multiple models.
The second group of methods represents the view of an entire model, which visualizes all extracted features of a model without highlighting decision-related information, such as Vanilla Backpropogation\cite{simonyan2013deep}, Guided Backpropogation\cite{springenberg2014striving}, and Deconvolution\cite{zeiler2014visualizing}.
This group of explanations' primary processing method is the backward pass, which is time-efficient and can produce fine-grained results.
However, this group of methods fails to explain models' decision-making convincingly because they indistinctively represent all the extracted information.

The third group of methods is the class-discriminative visual explanation \cite{zhou2016learning,selvaraju2017grad,fong2017interpretable,chattopadhay2018grad,omeiza2019smooth,wang2019score}, which can explain the model decision by localizing the regions essential for model predictions and is sensitive to different classes.
Zhou et al. \cite{zhou2016learning} introduce CAM (Class Activation Map), which is an initial approach of localizing a specific image region for a given image class.
However, researchers have to re-train the entire model to get the results of CAM.
As an approach to address the shortcomings of CAM, Grad-CAM \cite{selvaraju2017grad} is proposed as a more efficient approach, which can explain the predictions of CNN models without re-training or changing their structure.
In 2017, a perturbation-based method called BBMP \cite{fong2017interpretable} was introduced, which highlights the ROI of input images with the help of perturbations on input images.
Since BBMP requires additional pre-processing and multiple iterations, it was time-consuming and challenging to be implemented in real-time applications.
Recently, plenty of Grad-CAM-inspired methods have been proposed including Grad-CAM++\cite{chattopadhay2018grad}, Smooth Grad-CAM++\cite{omeiza2019smooth}, and Score-CAM\cite{wang2019score}.
Consistent with Grad-CAM, these methods are applicable to a wide variety of CNN models.

Aiming to provide interpretable CNN model comparison, we include multiple class-discriminative visual explanation methods to support the understanding of models' decisions.

% Model Comparison Works
\subsection{CNN Model Comparison}
Lots of previous works are aiming to address the need for CNN model comparison.
To assist researchers in CNN model evaluation and comparison, Canziani et al. \cite{canziani2016analysis} develop a quantitative analysis of fourteen different CNN models based on the accuracy, memory footprint, number of parameters, operations count, inference time, and power consumption. 
In terms of statistical analysis, multiple findings concerning the relationships of model parameters are discussed in \cite{canziani2016analysis}, such as the independence between power consumption and architecture, the hyperbolic relationship of accuracy, and inference time.
Liu et al. \cite{liu2017survey} go through four kinds of deep learning architectures, including autoencoder, CNN, deep belief network, and restricted Boltzmann machine.
It also illustrates those architectures' applications in some selected areas such as speech recognition, pattern recognition, and computer vision.  
A recent survey by Khan et al. \cite{khan2019survey} discusses the architecture development of deep CNNs, from LeNet\cite{lecun1995learning} presented in 1998 to Comprehensive SqueezeNet\cite{hu2018competitive} presented in 2018.
\cite{khan2019survey} offers a detailed quantitative analysis of twenty-four deep CNN models, comparing information such as the number of parameters, error rate, and model depth.

Besides these general model comparison studies \cite{canziani2016analysis,liu2017survey,khan2019survey}, researchers also apply the comparative study of multiple models for specific tasks. 
Aydogdu et al. \cite{aydogdu2017comparison} quantitatively compare three different CNN architectures based on their performance in the age classification task. 
Talebi et al. \cite{talebi2018nima} train multiple models to automatically assess image quality and compare their performances based on the accuracy and other quantitative measurements. 
Mukhopadhyay et al. \cite{mukhopadhyay2019performance} apply the performance comparison of three CNN models for the Indian Road Dataset, which represents the road detection results through images and compares the models based on the detection accuracy. 
By discussing past model comparison studies, we conclude that conventional works focus on quantitative comparison or structure analysis, which fail to reveal the underlying reasons for models' performances. 
To fill this research gap, our system would integrate XAI techniques, specifically, the visual explanation methods, to help researchers compare the deep CNN models in an interpretable way.

\subsection{Visual Analytics for Interpretable CNN Comparison}
A variety of visual analytics tools are aiming at supporting interpretable comparisons of CNN.
Some focus on visualizing and interpreting the internal working mechanism of a single CNN model \cite{rauber2016visualizing, liu2016towards, wongsuphasawat2017visualizing, bilal2017convolutional, kahng2017cti, hohman2019s, schubert2020openai}, combining various visualization techniques such as dimension reduction for understanding networks' hidden activities \cite{rauber2016visualizing}, \rev{a directed acyclic graph to disclose multiple neurons' facets and interactions \cite{liu2016towards}}, hierarchy analysis of similar classes \cite{bilal2017convolutional}, or feature visualizations and interactions \cite{hohman2019s}.
However, such in-depth inspection of a single model helps develop interpretation but is insufficient for scenarios where model comparison and selection are needed.

Researchers have developed some visual analytics frameworks for comparing CNN models \cite{krause2016interacting, zeng2017cnncomparator, zhang2018manifold, das2019beames, murugesan2019deepcompare, arendt2020parallel, ma2020visual, li2020visual}.
Prospector \cite{krause2016interacting} leverages partial dependence plots to visualize different performances of multiple models on one input sample.
To assist model training, CNN Comparator \cite{zeng2017cnncomparator} compares models from different training stages \rev{in the aspect of} model structures, parameter distributions, etc.
Utilizing predictions of the labels, Manifold \cite{zhang2018manifold} allows users to compare multiple models at the feature level using scatter plots.
BEAMES \cite{das2019beames} is a multimodel steering system providing multi-dimensional inspection to help domain experts in model selection.
\rev{However, these methods lack interpretability because they mainly use numerical features of CNNs.}
To assist interpretable comparison, \rev{researchers apply techniques such as linking model structures instances for comparing two binary classifiers \cite{murugesan2019deepcompare}, visualizing qualitative differences in how models interpret input data \cite{arendt2020parallel}, etc.
These techniques can assist better model interpretation, but only support the comparison among a small number of models.}
% DeepCompare \cite{murugesan2019deepcompare} presents a framework to link model structures and test instances of two binary classifiers.
% Parallel Embeddings \cite{arendt2020parallel} applies concept-oriented visualizations to convey semantic differences between models, helping data scientists understand how models interpret data.

In conclusion, most of the existing visual analytics methods for interpretable CNN comparison are either based on handcrafted quantitative parameters or only supporting comparison between two models. \rev{Only few of them allow CNN interpretation and multi-model comparison at the same time.}
\rev{With comparing and interpreting different CNN models becoming a growing demand,} there is a need for comparative studies that support a larger number of models' comparisons and present quantitative and qualitative information at the same time for more thorough evaluations.

\section{Design Goals}
\label{sec:DesignGoals}
\begin{figure*}[ht]
  \centering
  \includegraphics[width=0.85\textwidth]{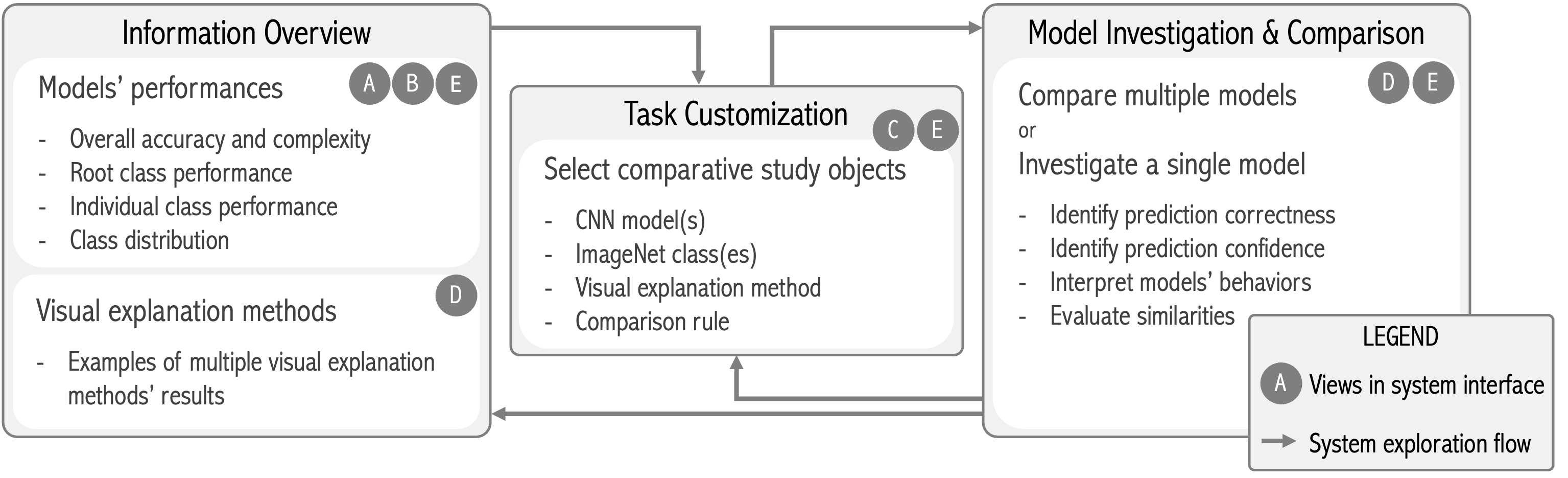}
  \caption{
  The comparative analysis workflow of VAC-CNN consists of three phases: information overview, task customization, and model investigation \& comparison. Each phase involves user interaction with multiple views of our system.
  }
  \label{fig:workflow}
\end{figure*}

According to our survey, we are aware of the need for CNN comparison tools that support flexible customization of comparative tasks (e.g., the in-depth inspection of a single model and comparative studies of multiple models).
Such tools should also integrate XAI techniques to assist model interpretation.
We refine this requirement into four design goals and describe them as follows.

\begin{compactenum}
  \item[\textbf{G1}] \textbf{Novice-Friendly Information Overview:}
    Motivated by the superb learning power of CNN models, researchers from different domains with various knowledge backgrounds are attempting to take advantage of this fast-developing technique in recent years \cite{li2021survey}.
    A visual analytics system for CNN comparison can be helpful for beginners as well as experts to gain more insights on models' behaviors.
    % Given that most of the existing model comparison tools are developed for experienced ML researchers, our system needs to provide a high-level information overview to help users build a general understanding of model performances and the XAI techniques we integrated.
    Given that most of the existing model comparison tools are developed for experienced ML researchers, our system needs to provide \rev{an information overview that can assist users in a high-level model screening based on their performances and a general understanding of the XAI techniques we integrated.}
    Moreover, the system should distill information and enable interactions to assist the overview process instead of overwhelming users with too many details all at once.
  \item[\textbf{G2}] \textbf{Informative Visual Explanation:}
    The commonly employed visual explanation methods based on color heatmaps highlighting the associated ROI are shown to be helpful in interpreting  CNNs~\cite{zhou2016learning,selvaraju2017grad,fong2017interpretable,chattopadhay2018grad,omeiza2019smooth,wang2019score}.
    However, it is hard to efficiently identify differences among models only based on the qualitative results from such visual explanation methods in the model comparison scenarios. 
    Thus, we need to consolidate the visual explanation methods with quantification measurements
    %(e.g., contour lines) 
    to help users gain better insights during the comparison process.
    Besides, when localization is not enough for interpreting a model, our system should provide complementary visualization for further analysis and help users better understand the underlying reason behind the CNN model predictions.

%   \item[\textbf{G3}] \textbf{Flexible Comparison Customization:}
  \item[\textbf{G3}] \textbf{\rev{Scalability and Flexibility:}}
    Unlike ML experts, beginners without comprehensive ML knowledge can benefit from additional exploration in a broader range of models when comparing models.
    % Therefore, it can be helpful if a visual analytics tool supports customization of model comparison tasks and accepts a flexible number of models \cite{zhang2018manifold}.
    % Therefore, it can be helpful if a visual analytics tool \rev{supports scalability to the number of models and accepts comparison task customization over a flexible number of models \cite{zhang2018manifold}.}
    Therefore, \rev{they need a a visual analytics tool that supports scalability in the number of models to compare and flexibility in the customization of comparison tasks \cite{zhang2018manifold}.}
    However, most of the existing comparison approaches for analyzing model behaviors only focus on two-model comparison \cite{murugesan2019deepcompare, zeng2017cnncomparator}.
    % To fill in this gap, we need to support flexible customization of CNN comparison tasks in our system and allow users to customize objects such as the number of the model(s), data class(es), and the visual explanation method(s). 
    To fill in this gap, we need to support \rev{scalable and flexible} CNN comparison tasks in our system and allow users to customize objects such as the number of the model(s), data class(es), and the visual explanation method(s). 

  \item[\textbf{G4}] \textbf{Real-time Interaction:}
    It could take a tremendous amount of GPU time to generate models' visual explanation results \cite{fong2017interpretable, wang2019score}, especially for large-scale datasets.
    With a web-based approach, we expect our system to be efficient enough to offer users a responsive interface, which means users should not experience a noticeable delay when exploring model comparison scenarios through our system.
    Besides, we should allow users to interactively audit details of each view and select specific elements to inspect further information.
    Moreover, it is essential to present multiple views synergistically and help users better understand the models through the coordinated information of each view.
\end{compactenum}

\section{Methodology}
\label{sec:Methodology}
VAC-CNN is built upon thirteen \rev{widely-used} models, \rev{to cover various state-of-the-art architectures} such as AlexNet\cite{krizhevsky2012imagenet}, ResNet \cite{he2016deep}, SqueezeNet \cite{iandola2016squeezenet}, DenseNet \cite{huang2017densely}, MobileNet \cite{howard2017mobilenets}, and ShuffleNet \cite{zhang2018shufflenet}.
The models are pre-trained on the ImageNet dataset\cite{russakovsky2015imagenet} for the image classification task and we develop our system on the ImageNet (ILSVRC2012) validation set with $1,000$ image classes and $50,000$ images.
In this section, we introduce the analysis workflow and the integrated methodologies of our system.

\subsection{Workflow}
Based on our survey of existing tools \cite{krause2016interacting, zeng2017cnncomparator, zhang2018manifold, das2019beames, murugesan2019deepcompare, arendt2020parallel, ma2020visual, li2020visual} and the design goals in Sec. 3, we model the comparative analysis procedure with VAC-CNN as a three-phase workflow (see Fig.\ref{fig:workflow}).
The workflow starts from \textbf{Phase 1} which provides an information overview to help ML beginners get a general understanding of both model performances and visual explanation methods. 
\textbf{Phase 2} provides task customization to support flexible study options towards CNNs, ImageNet classes, visual explanation method, and comparison rule.
Based on the customized comparison requirements, \textbf{Phase 3} presents coordinated visualizations and qualitative information for multi-model comparison or single-model investigation, respectively. 
We will connect our discussion about the methodology in this section and the interface design in the following section with these phases.

\subsection{Distribution Graph Generation}

In regard to design goal \textbf{G1}, we provide a comprehensive and novice-friendly information overview for our users to understand the model's high-level performances.
One way to achieve this is to investigate the class distribution which is generated from the model's prediction and reflects how the model interprets the data.

To visualize the distribution of image classes in respect to a specific model, we create a distribution graph (Fig.~\ref{fig:interface} (B) based on each model's predictions.
In Algorithm \ref{alg:DistanceM}, we use confidence matrix as the baseline to generate this graph, since it reflects how a model understands the input data.
For a given model, every input image is classified according to a confidence vector of size $(1, N)$, where $N$ is the number of image classes.
The confidence vector is generated from the Softmax function and presents the model's prediction of the input.
By concatenating the confidence vectors of all input images, we get the model's confidence matrix of size $(M, N)$, which includes the model's predictions of the entire input dataset of size $M$.

\begin{algorithm}[hbt]
\caption{Constructing Distance Matrix of $N$ ImageNet Classes}
\label{alg:DistanceM}
\begin{algorithmic}[1]
\REQUIRE ~~\\
The image class list of all images in the dataset, $imgClasses$;\\
The confidence matrix of the model, $confMat$;
\ENSURE ~~  \\
The distance matrix of $N$ image classes, $distMat$;
\STATE{$distMat$ $\gets$ $zeros((N, N))$}
\STATE{$distMatCount$ $\gets$ $zeros((N, N))$}
\STATE{\textcolor{gray}{// iterate through all images}}
\FOR{$imgIdx$ \textbf{in} $range(M)$}
\STATE{$curClass$ $\gets$ $imgClasses[imgIdx]$ \ \textcolor{gray}{// get ground-truth class}} 
\STATE{$P$ $\gets$ $confMat$[$imgIdx$, $:$]  \ \ \ \ \ \textcolor{gray}{// get confidence score vector}}
\STATE{\textcolor{gray}{// iterate through all classes}}
\FOR{$compClass$ \textbf{in} $range(N)$} 
\STATE{$confScore$ $\gets$ $P[compClass]$}
\STATE{$distMatCount[curClass, compClass]$ $\mathrel{{+}{=}}$ $1$}
\STATE{\textcolor{gray}{// calculate distance}}
\IF{$compClass$ $\neq$ $curClass$}
\STATE{$distMat[curClass, compClass]$ $\mathrel{{+}{=}}$ $(1 - confScore)$}
\ENDIF
\ENDFOR
\ENDFOR
\STATE{$distMat$ $\gets$ $divide(distMat$, $distMatCount)$ \textcolor{gray}{// get the average}}
\STATE{$distMat$ $\gets$ $[distMat + transpose(distMat)] / 2$}
\end{algorithmic}
\end{algorithm}

Based on the confidence matrix, the distance matrix $distMat$ of the $N$ image classes is generated as described in Algorithm \ref{alg:DistanceM}.
Firstly, the distance matrix $distMat$ of the $N$ image classes is initialized as a zero matrix of size $(N, N)$.
Then we assign the class of each input image to $curClass$ and each image's confidence matrix of size $(1, N)$ to $P$ (line 4-5).
After that, we iterate over the $distMat$ and update the value through the iteration among $P$ (line 8-14).
Then, we calculate the $distMat$ using the iteration results and the iteration counts $distMatCount$ (line 16-17).
Finally, we apply dimensionality reduction to the resulting matrix $distMat$ using t-SNE \cite{maaten2008visualizing} to generate the 2D projection matrix for the distribution graph.

The distribution graph presents the distribution of the $N$ image classes with respect to the predictions of each CNN model.
Since the ImageNet structure is based on the WordNet hierarchy, there are eight root classes representing how human beings classify the $N=1000$ image classes in the ImageNet dataset.
In the distribution graph of our system, each root class is represented by a specific color, allowing users to compare the model's classification and human's classification easily.

\subsection{Visual Explanation Methods}

% As discussed in Sec. 2.3, XAI techniques can be helpful for novice ML practitioners to understand model behaviors.
As discussed in Sec. 2.3, \rev{visual explanation methods, especially the class-discriminative ones, can help} novice ML practitioners to understand CNN model behaviors.
\begin{figure}[b]
  \centering
  \includegraphics[width=8cm]{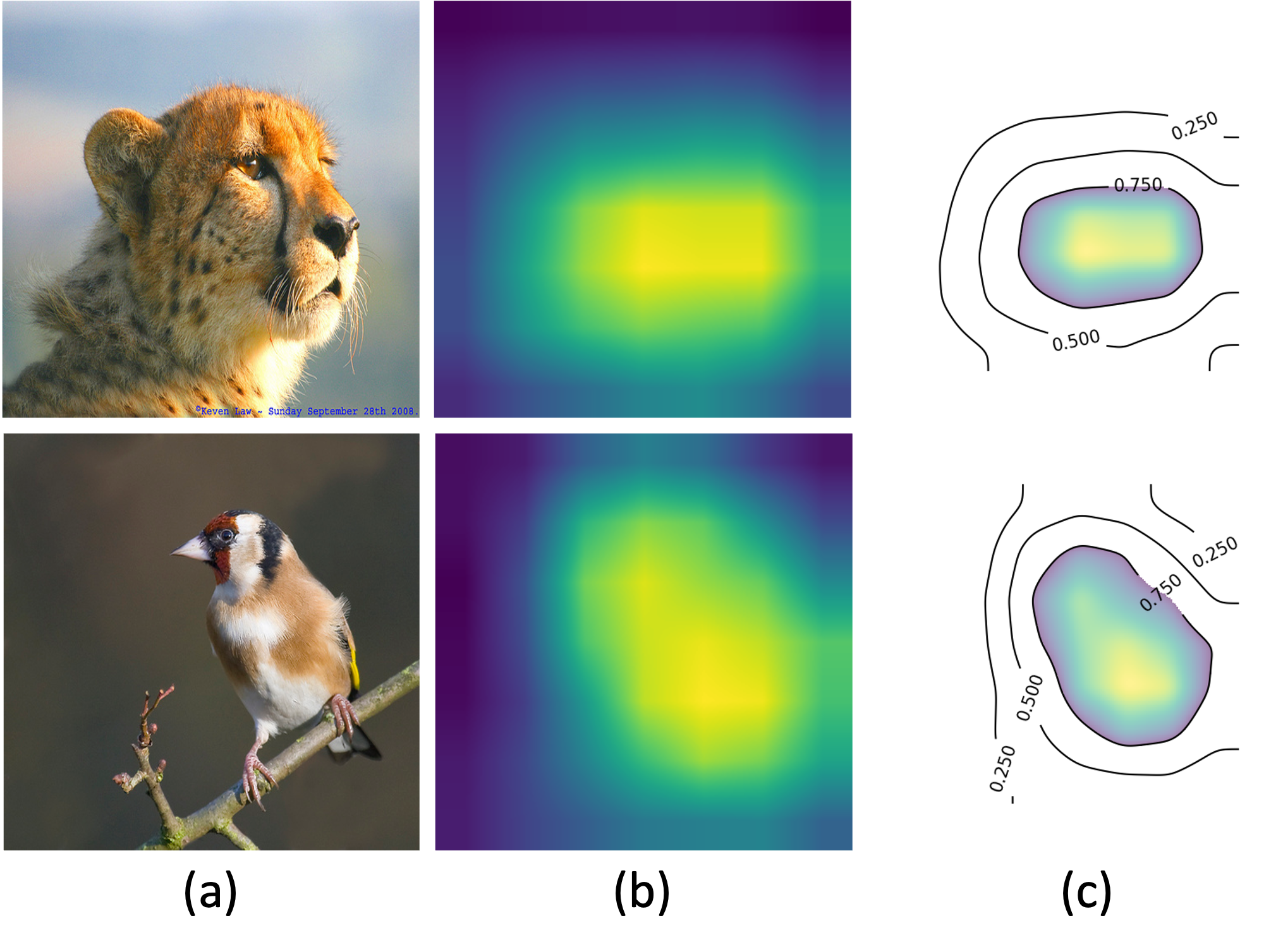}
  \caption{
  Examples of how we improve the presentation of the visual explanation method's results in VAC-CNN. Images at the same row are associate with the same original image. From left to right: \textbf{(a)} The original images; \textbf{(b)} The conventional visualizations of a visual explanation method; \textbf{(c)} The improved visualizations in VAC-CNN.
  }
  \label{fig:contour}
\end{figure}
\rev{Because they can highlight specific regions on the input image that is inferred to contribute the most to the model's decision-making.}
However, most of the existing visual analytics tools for CNN comparison do not include any visual explanation methods.
To fill this gap, we include five class-discriminative visual explanation methods in VAC-CNN, \rev{including} Grad-CAM \cite{selvaraju2017grad}, BBMP \cite{fong2017interpretable}, Grad-CAM++ \cite{chattopadhay2018grad}, Smooth Grad-CAM++ \cite{omeiza2019smooth}, and Score-CAM \cite{wang2019score}.
\rev{Examples of these five methods are shown in Fig. \ref{fig:interface} (D).}
\rev{The reason why these five methods are included is to cover} multiple kinds of methods such as gradient-based explanations (Grad-CAM, Grad-CAM++, Smooth Grad-CAM++), perturbation-based explanation (BBMP), and score-based explanation (Score-CAM), \rev{which supports our design goal \textbf{G2}}.
Our analytics system is designed to be extensible, so other visual explanation methods can be easily added.

To achieve our design goal \textbf{G2}, we consolidate the presentation of the visual explanation method's result.
As shown in Fig. \ref{fig:contour} (b), a conventional approach to present the visual explanation method's results is showing the heatmap, which doesn't provide any direct quantitative information.
Thus, the subtle difference among multiple heatmaps can be hard to identify, making it not informative enough for the model comparison task.

In VAC-CNN, we add the quantitative information about the visual explanation method's result by overlaying multiple contour lines over the heatmaps \rev{\cite{hunter2007matplotlib}}, which are associated with the attention matrix generated by the visual explanation method \rev{(with attention scores of $[0, 1]$, $0$ for ``no attention")}.
To support the highlighting of ROI, we also add a customizable threshold for users to remove regions of little attention accordingly.
\rev{For example, a threshold of $0.5$ means the region with attention scores lower than $0.5$ will not be highlighted.}
As shown in Fig. \ref{fig:contour}(c), our improved visualizations of the explanation results incorporate qualitative information and quantitative measures of the attention level, which can support users in model comparison tasks more effectively.

\subsection{Similarity Matrix Generation}

When comparing multiple models based on a single image, users can benefit from a  similarity matrix that intuitively shows the correlation of visual explanation methods' results for the CNN models.
We demonstrate the method to construct such similarity matrix in Algorithm \ref{alg:SimilarityM}.
\begin{algorithm}[b]
\caption{Constructing Similarity Matrix of Selected Models}
\label{alg:SimilarityM}
\begin{algorithmic}[1]
\REQUIRE ~~\\
The list of visual explanation results of models, $expResults$;\\
The function of computing similarity scores, $simFunc$;
\ENSURE ~~  \\
The similarity matrix of selected models, $simMatrix$;
\STATE{$L$ $\gets$ $len(expResults)$}
\STATE{$simMatrix$ $\gets$ $zeros((L, L))$}
\FOR{$idx1$ \textbf{in} $range(len(expResults))$}
\FOR{$idx2$ \textbf{in} $range(len(expResults))$}
\STATE{\textcolor{gray}{// iterate through all visual explanation results}}
\STATE{$expRes1$ $\gets$ $expResults[idx1]$}
\STATE{$expRes2$ $\gets$ $expResults[idx2]$}
\STATE{$simMatrix[idx1, idx2]$ $\gets$ $simFunc(expRes1, expRes2)$}
\ENDFOR
\ENDFOR
\end{algorithmic}
\end{algorithm}

In this algorithm, the generated saliency map from visual explanation methods are stored as matrices in a list \textit{expResults}.
\rev{We provide multiple widely-used image similarity measurements, including the structural similarity index (SSIM), the mean-square} \rev{error (MSE), the $L1$ measure, and the hash function.}
\rev{The default} \rev{similarity measurement is set to $L1$ because of its wide acceptance, and other options are provided for users to select a different rule as needed.}
Based on the user-specified similarity comparison rules, we use the corresponding function \textit{simFunc} to calculate the similarity score of two visual explanation results.
After iterating over every element of \textit{expResults}, we can get the similarity matrix \textit{simMatrix} quantifying the similarity of each pair of the visual explanation results.
To represent the value intuitively, we use \textit{seaborn} \rev{\cite{waskom2021seaborn}} to generate the resulting matrix's heat map.
Then, users can interactively compare the behaviors of the selected CNN models through our designed interface described in Sec. 5.

\subsection{Image Statistical Analysis}
In some circumstances, the conventional visual explanation method may fail to provide enough information to explain the model's prediction.
For example, when the prediction result is wrong but the localization is correct,  visual explanation methods doesn't explain why the model made a wrong decision.
To solve this problem, we go one step further by analyzing the information generated from the image region that the model cites as essential.

\begin{figure}[b]
  \centering
  \includegraphics[width=8.5cm]{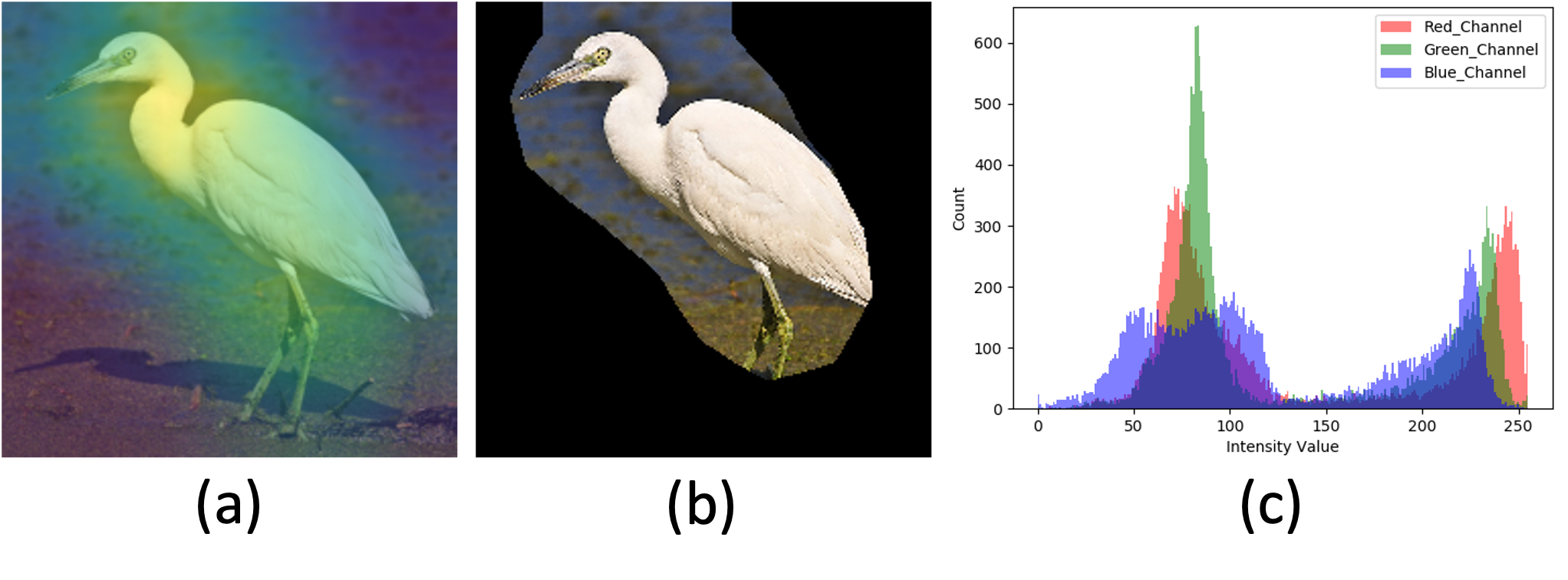}
  \caption{
    Example of the image statistical analysis process. From left to right: \textbf{(a)} A visual explanation method highlights ROI on the sample image; \textbf{(b)} The filtered image by only keeping the highlighted region of (a); \textbf{(c)} The color intensity histogram of the filtered image in (b).
  }
  \label{fig:histExample}
\end{figure}

As discussed in \cite{baker2018deep, geirhos2018imagenet}, CNN classifiers pre-trained on ImageNet have been proved to rely on texture information rather than the global object shape.
However, current algorithms using image texture are often deep-learning-based \cite{gatys2016image, cai2019diversity}, which can severely interfere our system's response speed.
In VAC-CNN, we apply color intensity histograms (CIHs) to measure image information, which are commonly used to analyze the image content and evaluate the image similarity \cite{roy2013image,malakar2013image,phetnuam2018classfication}.
In this way, the analysis results can be generated in real-time (\textbf{G4}).
Our process of image statistical analysis is shown in Fig. \ref{fig:histExample}.
Based on the model's explanation, which highlights a specific image region essential for the model to make predictions, we can filter the original image by removing the ``inessential'' part.
Then we visualize color intensity information of the filtered image (Fig. \ref{fig:histExample} (C)) to depict the statistical details of the image region that the model cites as essential in making predictions.

As the supplementary information of the visual explanation result, the color intensity histogram can help users further analyze what information the model extracts from the input.
Through comparing the visual explanation results and the color intensity histograms, users can gain more insights into the underlying behaviors of the deep CNN model.

\section{System Interface}
\label{sec:SystemInterface}
To achieve our design goals described in Sec. 3, we integrate the techniques introduced in Sec. 4 into a web-based visual analytics system, VAC-CNN, for the comparative studies of deep CNN models.
As shown in Fig. \ref{fig:interface}, the system interface includes five primary views: ``Overall Information View'' (A), ``Distribution Graph View'' (B), ``Task Selection Sidebar'' (C), ``Visual Explanation View'' (D), and ``Supplemental View'' (E). 
In this section, we illustrate how these views coordinate to facilitate the three phases of comparison workflow described in Sec. 4.1.

\subsection{Information Overview}
In order to assist non-experts ML practitioners (\textbf{G1}), VAC-CNN provides an information overview for users to explore high-level CNN model performances and the general behaviors of multiple visual explanation methods.
The analysis in \rev{this} phase requires information from View (A), (B), (D), and (E) in our visual interface.

The Overall Information View (Fig. \ref{fig:interface} (A)) illustrates the overall and detailed quantitative information of the included CNN models with multiple visualizations.
The scatterplot labeled as (A1) indicates each model's complexity and overall accuracy on the entire ImageNet validation set, where each point represents a CNN model.
% By clicking a point, users can pin a model and update the following zoomable bar charts (A3) and (A4) to investigate performance details concerning the selected model.
The radar chart labeled as (A2) reveals the accuracy performance of \rev{the selected models} on the eight root classes.
Each line of the radar chart represents one model's performance, and the selectable legend located at the right of the chart enables users to \rev{remove uninterested models and only compare selected ones}.
% Additionally, our interactive design allows users to change the pinned model or the pinned root class by clicking the line or the root class name of the radar chart, which can update bar charts shown at (A3) and (A4) of Fig. \ref{fig:interface}.
Additionally, \rev{our interactive design allows users to change the pinned model or the pinned root class by a simple click, which can update the two zoomable bar charts shown at (A3) and (A4) of Fig. \ref{fig:interface}, representing leaf classes' accuracy information of the model and root class, respectively}, where the leaf classes are ranked in descending order of their accuracies.
% Thus, each part of the Overall Information View can work synergistically to illustrate each CNN model's quantitative information from multiple aspects, helping users develop high-level knowledge of models' performances.
Thus, each part of the Overall Information View can work synergistically to illustrate each CNN model's quantitative information from multiple aspects, helping users \rev{perceive models' performances and achieve efficient high-level multi-model screening accordingly}.

The Distribution Graph View (Fig. \ref{fig:interface} (B)) reveals the distribution of the $1000$ ImageNet classes.
Each point represents a single image class, and the colors correspond to eight root classes.
Generated from each model's confidence score matrix, this visualization presents the model's class-level behavior, enabling users to discover the model's coherent or inconsistent behaviors across clusters of image classes.
Besides, by looking at the clusters, users can also discover typical image class groups for further investigation in the following phases, which means this view also serves as a class recommendation.
Similarly, smooth user interactions, including hovering over, clicking, zooming, etc., are supported as well.

\rev{As discussed in Sec. 4.3,} the Visual Explanation View (Fig. \ref{fig:interface} (D)) presents the example results of multiple visual explanation methods, informing non-expert users how each of the visual explanation method's result looks like.

Finally, the Supplemental View (Fig. \ref{fig:interface} (E)) provides users with supplemental information.
At the information overview phase, two bar charts are presented at this view before users make any ImageNet class selections at the Task Selection Sidebar (Fig. \ref{fig:interface} (C)).
The first bar chart, ``Range of Class Accuracy'', visualizes the range of the thirteen models' accuracies on six image classes, including image classes on which the models have either diverging or parallel performances.
And the second bar chart, ``Average of Class Accuracy'', includes information related to six image classes, on which the models have coherent good performances or bad performances.
These two bar charts illustrate image classes with abnormal statistical characteristics, suggesting interesting images for users to explore in more detail.

\subsection{Task Customization}
VAC-CNN also supports users to customize the comparative study (\textbf{G3}) with the Task Selection Sidebar at the bottom left of our system interface (Fig. \ref{fig:interface} (C)).
From this view, users can select multiple CNN model(s), ImageNet class(es), visual explanation method(s), etc.
Based on different selections, multiple subtasks can be performed in the following phase, including comparing multiple models over a particular image class, investigating a single model's behaviors on multiple image classes, and explaining a single model's behavior on images within a particular class.
\rev{For the multi-model comparison task, VAC-CNN supports the users to select up to $13$ models for comparison.}

\subsection{Model Investigation \& Comparison}
In the model investigation \& comparison phase, the views (D) and (E) will be updated to present information based on the result from the user-specified comparison task (\textbf{G3}).

In the Visual Explanation View, various information is presented through a table format representation \rev{to better achieve the design goal \textbf{G3}}.
\rev{With multiple rows, this table presents comparison results of up to $13$ models selected by users through task customization (as described in Sec. 5.2).}
\rev{Besides, the interaction features allow users to sort on multiple quantitative columns} and search specific information to filter the results and get a deeper understanding.

\rev{We present an example to demonstrate what information is presented in this table.}
For instance, in the single-model investigation task described in Sec. 6.2, the view presents information including: 
\begin{itemize}
\setlength\itemsep{-0.2em}
\item the quantitative performance measures, such as model's overall accuracy, class accuracy, confidence score, etc.; 
\item the corresponding information useful for understanding and comparing, such as the model name, image's ground-truth class, and predicted class;
\item the visual explanation results presented as contour plots, explanations on original images, as well as the CIH for the highlighted image region, etc.
\end{itemize}

\begin{figure*}[t]
\centering
\includegraphics[width=1\textwidth]{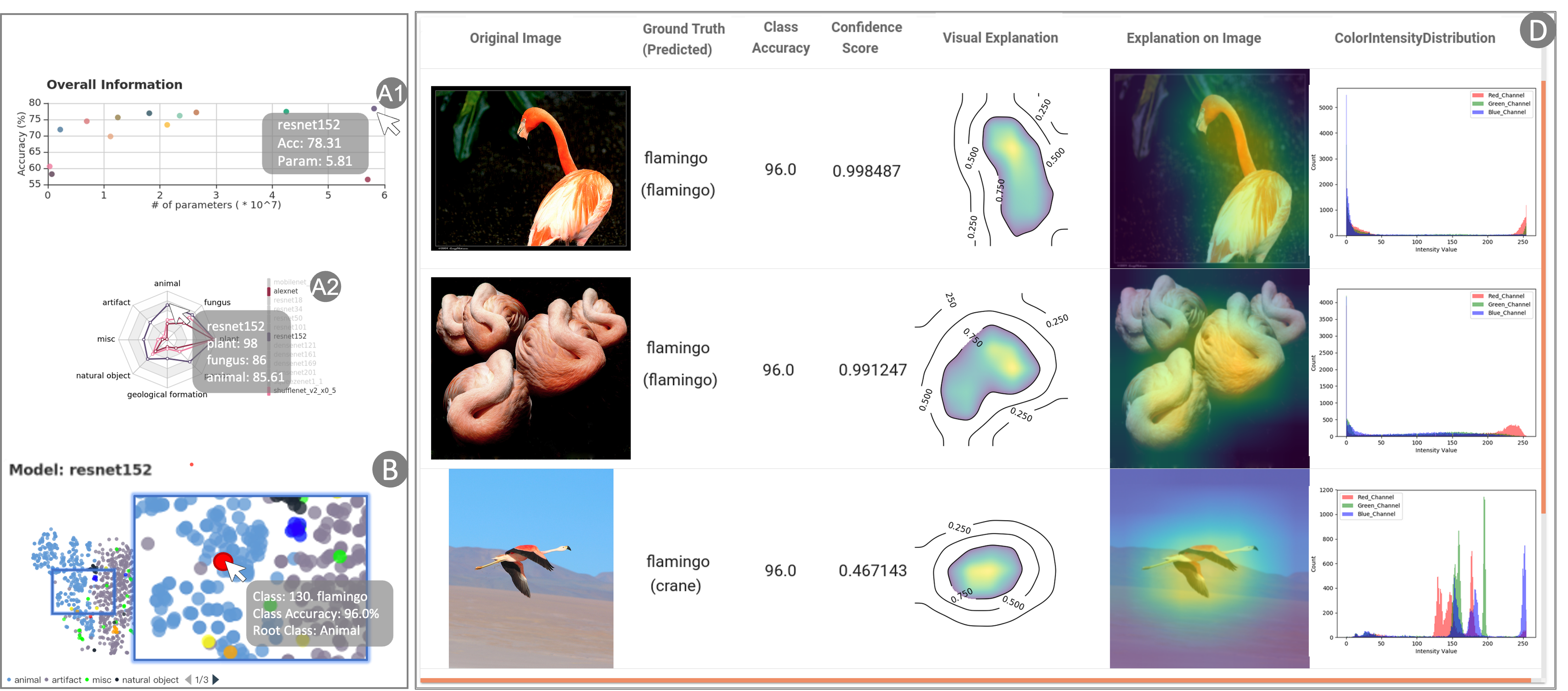}
\caption{
Investigating a single model \textit{resnet152} on different images of class ``flamingo'' using VAC-CNN (Sec. 6.2).
The multiple visualizations in (A1), (A2), and (B) can help users notice the high-level behavior pattern of \textit{resnet152} from different perspectives.
And the coordinated view (D) shows the CIH results to help users discover the common patterns among models' failure cases on the ``flamingo'' class.
(The views are marked as (A1), (A2), (B), (D) to keep consistent with Fig. \ref{fig:interface}.)
}
\label{fig:rgbHist}
\end{figure*}

Specifically, the CIH is used for supporting the single-model investigation tasks, so VAC-CNN only presents CIH when users are investigating a single model, as shown in Fig. \ref{fig:rgbHist} (D).
\rev{As discussed in Sec. 4.3}, VAC-CNN enables threshold adjustment for users to update the threshold of contour visualization of the visual explanation results.
VAC-CNN coordinates the above information to support the comparative study process (\textbf{G2}).

\begin{figure*}[t]
\centering
\includegraphics[width=1\textwidth]{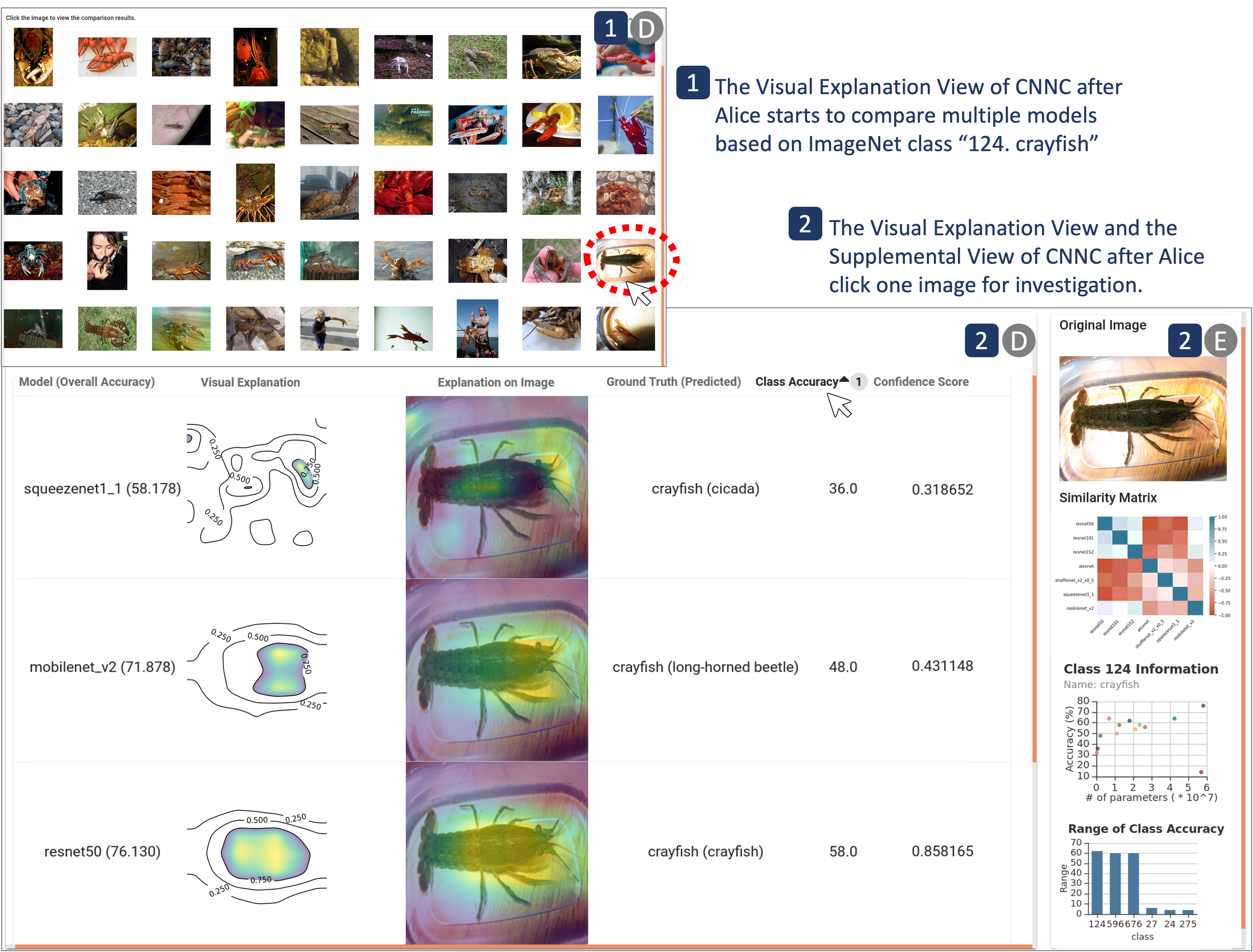}
\caption{
% Comparing multiple models on a selected image from class ``\textit{$124$ crayfish}'' using VAC-CNN (Sec. 6.1).
Comparing \rev{the performance of 7} models on a selected image from class ``\textit{$124$ crayfish}'' using VAC-CNN (Sec. 6.1).
Looking over in (1-D), we can find a common characteristic of images in this class is their complicated background.
% Looking over (1-D), users discover one typical pattern of the original images in this class is the complicated image background.
After selecting one image \rev{of} a shrimp, we can find more information with the updated views (2-D) and (2-E) (The views are marked as (D) and (E) to keep consistent with Fig. \ref{fig:interface}).
By sorting the result in the table of (2-D) with different metrics, we can check the potential relationship between the model's prediction performances and the ROI size.
\rev{(Note: We compare 7 models in this task, but only show 3 models' result in (2-D) because of space limitation.}
}
\label{fig:exp1}
\end{figure*}

In the Supplemental View, users can find various supplementary information to support model comparison and investigation according to different analysis needs.
When users compare multiple models, this view includes information such as the original image selected by users, the similarity matrix of the models' visual explanation results, and the scatterplots presenting the models' accuracies on each selected image class (Fig. \ref{fig:exp1} (2-E)).
When users investigate a single model, this view only shows the accuracy scatterplots of each selected image class since most of the essential information is already available in the Visual Explanation View.

\section{Use Cases}
\label{sec:UseCases}

In this section, we demonstrate how VAC-CNN can help novice ML practitioners conduct comparative studies with two use cases: (1) comparing the behavior of multiple models on the same image, (2) investigating a single model's behavior on different images.
\rev{The first use case demonstrates how VAC-CNN supports multi-step model comparison, from high-level screening over $13$ models to the in-depth interpretable comparison of $7$ models.
The second use case is about single-model inspection, showing how our provided informative visual explanation assists users.}

\subsection{Comparing Multiple Models on Same Image}
Alice is a Master's student majoring in animal behavior study.
She gained some basic knowledge about CNN and deep learning from a public course provided by the Computer Science Department and wants to apply it to her own major.
Therefore, she uses VAC-CNN to explore the performances and behaviors of multiple CNN models on a group of images about animals.

After opening up the system, Alice starts by deciding the models, ImageNet classes, and visual explanation methods for her comparison task.
She looks over the different plots in the Overall Information View (Fig. \ref{fig:interface} (A)) \rev{to inspect the performances of the $13$ CNNs} and becomes interested in ResNet models when she notices the performance boost from \textit{resnet18} to \textit{resnet152}.
She also notices that ResNet architectures often have good performance on the ``animal'' group from the radar chart in this view.
Moreover, Alice finds ``animal'' forms better cluster for \textit{resnet50} from the Distribution Graph View (Fig. \ref{fig:interface} (B)), so she adds \textit{resnet50}, \textit{resnet101}, \textit{resnet152} into the list of models.
Then she looks at the first bar chart in the Supplemental View (Fig. \ref{fig:interface} (E)) and finds that the models' accuracies vary significantly on class ``\textit{$124$ crayfish}'', which belongs to the ``animal'' group, so she decides to choose this class for model comparison.
Finally, Alice explores the Visual Explanation View (Fig. \ref{fig:interface} (D)) and notices the ROI provided by ``Grad-CAM'' is very clear in general, so she decides to use ``Grad-CAM'' as the visual explanation method. 

After having the models, ImageNet classes, and visual explanation methods she wants to select in mind, Alice moves on to the Task Selection Sidebar to customize her comparison task (Fig. \ref{fig:interface} (C)).
\rev{When restricting} the ImageNet class selection to ``\textit{$124$ crayfish}'', Alice notices that a scatter plot in the Supplemental View is updated, as shown in Fig. \ref{fig:interface} (E).
% And two models with remarkably bad performance, \textit{alexnet} ($14\%$), \textit{mobilenet\_v2} ($48\%$), stands out.
\rev{And one model with remarkably bad performance, \textit{alexnet} ($14\%$), stands out.
Besides, there are $3$ other models whose accuracy is lower than $50\%$, \textit{shufflenet\_v2\_x0\_5} ($32\%$), \textit{squeezenet1\_1} ($36\%$), and \textit{mobilenet\_v2} ($48\%$).}
Curious about the reasons behind those models' failures, Alices also decides to add them to the model list for comparison.
In this way, Alice has finalized the objectives of the model comparison task \rev{with 7 models}: 
\begin{itemize}
    \item Models: \textit{resnet50}, \textit{resnet101}, \textit{resnet152}, \textit{alexnet}, \rev{\textit{shufflenet\_v2\_x0\_5}, \textit{squeezenet1\_1},} \textit{mobilenet\_v2};
    \item ImageNet Class: \textit{$124$ crayfish};
    \item  Visual Explanation Method: Grad-CAM.
\end{itemize}

After finishing all of the customizations, Alice starts the comparison with the Visual Explanation View (Fig. \ref{fig:interface} (D)) and the Supplemental View (Fig. \ref{fig:interface} (E)).
She first looks over the original images within the selected class (Fig. \ref{fig:exp1} (1-D)) from the Visual Explanation View.
She finds that the image background of the main object---``crayfish''---is very complicated for almost every image in this class, which can be a possible cause of the varied model performances.
With this hunch, Alice clicks on one image and begins to compare the models' behaviors with the updated Visual Explanation View and Supplemental View (Fig. \ref{fig:exp1} (2)).
As shown in Fig. \ref{fig:exp1} (2-D), by sorting the table according to the class accuracy, Alice inspects the visual explanation method's results and the associated numerical information of \rev{the $7$ models}.
She notices that\rev{, for the $3$ models shown in Fig. \ref{fig:exp1} (2-D),} \textit{resnet50} is the only model that correctly classifies the input, while both \rev{\textit{squeezenet1\_1}} and \textit{mobilenet\_v2} make incorrect predictions.
By inspecting the visual explanation methods' results, Alice realizes that the size of each model's ROI has a positive relationship to the model's prediction correctness: \textit{alexnet} (lowest class accuracy) only highlights a very small region while the ROI of \textit{resnet152} (highest class accuracy) is among the largest ones.
After checking more images, Alice confirms the consistency of this observation.
Given most of the images in this class have complicated backgrounds, Alice concludes that models with smaller views (i.e. smaller ROI) can't perform very well in this object classification task.
From this comparative study, Alice learns that when the images she is dealing with have complicated backgrounds, she will consider selecting CNNs with broader views (e.g., \textit{resnet50}) over others.

\subsection{Investigating a Single Model on Different Images}

This use case involves Bob, a first-year Ph.D. student majoring in Computer Science.
He is developing a bird recognition App for a course project.
And he wants to find the best model for the bird image classification function in his App.

Similar to Alice, Bob starts by deciding the model, ImageNet classes, and visual explanation methods for his task.
He first checks the models' differences in complexities and overall accuracies with the scatter plot in the Overall Information View (Fig. \ref{fig:rgbHist} (A1)).
He finds that \textit{resnet152} achieves the best performances compared to other CNN models.
And such an advantage is particularly prominent with the root classes ``animal'' and ``fungus'' according to the radar chart in the Overall Information View (Fig. \ref{fig:rgbHist} (A2)).
Therefore, Bob decides to choose \textit{resnet152} as the model to dig deeper into.
As shown in Fig. \ref{fig:rgbHist} (B), then he zooms into the distribution graph of \textit{resnet152} to check the cluster of ``bird'' species and decides to select class ``\textit{$130$ Flamingo}'' to explore model behaviors on it.
Finally, after looking over the examples of multiple visual explanation methods in the Visual Explanation View (Fig. \ref{fig:interface} (D)), he chooses ``Smooth Grad-CAM++'' as the visual explanation method for model interpretation.
In summary, the objectives of Bob's model comparison task are: 
\begin{itemize}
    \item Model: \textit{resnet152};
    \item ImageNet Class: \textit{$130$ Flamingo};
    \item  Visual Explanation Method: Smooth Grad-CAM++.
\end{itemize}

After customizing his comparison task with the Task Selection Sidebar, Bob starts the model investigation from the updated view (Fig. \ref{fig:rgbHist} (D)).
He first notices that Smooth Grad-CAM++ indicates correct localization of the main object in every image in the class ``\textit{$130$ Flamingo}'', even for those incorrectly predicted ones.
He feels excited about this discovery and continues to look for the cause of those incorrect predictions made by \textit{resnet152}.
He finds that \textit{resnet152} correctly classifies the first two images with high confidence scores but misclassifies the third as ``Crane'' in Fig. \ref{fig:rgbHist} (D).
In contrast, Bobs thinks the second image is more challenging to recognize than the third one in his eyes.
He tries to explain this phenomenon from the color intensity histograms (CIH) provided by VAC-CNN.
By comparing the CIHs of the three images, he realizes that the second image's CIH is highly similar to the first one, while the third image's CIH looks much more different from the other two (Fig. \ref{fig:rgbHist} (D)).
After checking the conditions with other image classes of bird species, Bob finds such observation still holds for most failure cases.
He shares this interesting discovery with his course instructor.
The instructor suggests he construct a small subgroup of the bird classes that most confuses \textit{resnet152}, apply data augmentation specifically, and use it to fine-tune the model.
Bob optimizes his model following this idea, and makes his bird recognition App more potent in the classification task.

\section{Preliminary Evaluation Study}
\label{sec:PreliminaryEvaluationStudy}
VAC-CNN is designed for assisting novice ML practitioners in comparing and understanding multiple CNN models.
In this section, we conducted a preliminary evaluation study to demonstrate the usefulness of our system.
Specifically, we intended to understand whether VAC-CNN was effective in helping users: (1) gain high-level understanding of various CNN models (\textbf{G1}); (2) interpret CNN behaviors (\textbf{G2}); (3) customize different comparison tasks (\textbf{G3}).
We also investigated them about how they felt about the smootheness of the system as well as the interactions (\textbf{G4}).
\rev{The evaluation of our study mainly adopts qualitative analysis towards participants' behaviors and feedbacks, along with minor quantitative analysis of their self-reported ML knowledge level and rating scores of the system.}

Considering the unprecedented challenging situation brought by Covid-19, our study environment was restricted and we had to do everything remotely with limited number of participants.
However, because we carefully designed the entire study procedure and address a thorough evaluation, the validity of VAC-CNN can still be proved through this study.

\subsection{Participants}
We recruited $12$ participants ($6$ male, $6$ female), including $7$ M.S. students and $5$ Ph.D. students.
We asked them to self-report their familiarity with three areas on a scale of $[0, 10]$ ($0$ for ``no knowledge" and $10$ for ``expert") and report the statistics as follows:
\begin{itemize}
\setlength\itemsep{-0.2em}
\item Basic machine learning techniques: $Md=4.00$, $IQR=2.25$; % (average score: 4); 
\item CNNs: $Md=2.50$, $IQR=1.25$; % (average score: 2.8); 
\item Visual explanation methods: $Md=2.00$, $IQR=1.25$. %(average score: 2.3)
\end{itemize}
The result shows that all of the participants have limited deep learning and XAI background, so they belong to our target user group, novice ML practitioners.

\subsection{Task Design}
We asked each participants to perform the same tasks using VAC-CNN and observe their behavior patterns during the process.
After getting familiar with the visual interface, they were asked to perform the following tasks:

\begin{compactenum}
  \item[\textbf{T1}] \textbf{Browse high-level information:}
    The participants were asked to get a high-level understanding of model performances and the behaviors of multiple visual explanation methods (\textbf{G1}) through interactions with multiple visualizations presented in our visual interface (\textbf{G4}).
    They were encouraged to use as much interactions as possible and describe their findings.

  \item[\textbf{T2}] \textbf{Compare multiple models:}
    To observe how VAC-CNN can assist users in multi-model comparison, we asked the participants to compare at least two models (\textbf{G2, G3}).
    The models, as well as other customizable options, such as visual explanation methods, were chosen by the participants.
    And we asked them to provide the reason of their selections.
    The participants were asked to identify common and unique behaviors of the compared models, and which components of VAC-CNN lead to their findings.
    
  \item[\textbf{T3}] \textbf{Investigate a single model:}
    In this task, the participants were asked to select one CNN model for in-depth investigation.
sure    Similar with task T2, we asked them to decide all customizable options, including the model they chose to investigate, and provide us with the reasons (\textbf{G3}).
    The participants were asked to describe their understanding of model behaviors and how VAC-CNN assist them during the process (\textbf{G2, G4}).
\end{compactenum}

\subsection{Study Setup and Procedure}
Our preliminary evaluation study is conducted remotely through one-on-one video meeting with each participant.
The participants were asked to access VAC-CNN running at a remote server with their personal computers.
Before the study started, we asked each participant to self-report their knowledge background and basic demographic information.
At the beginning of the study, we provided a 5-minute tutorial session to introduce the models, dataset, visual components, and interactions built in VAC-CNN.
After that, we asked the participants to perform the three tasks described in Sect. 7.2, and encourage them to use as many system components as possible.
This session took around $30$ minutes on average and participants followed the think-aloud protocol when they performed these tasks.
Finally, the participants were invited to fill up a usability questionnaire and share their feedback about experiences with VAC-CNN in a 5-minute follow-up interview.

\subsection{Results and Discussion}
\rev{This section demonstrates our findings from the usability questionnaire, the follow-up interview, and the behavior observation of all users.}
We asked the participants to rate the usability of the system \rev{in the questionnaire} as well as collected their comments about the system in a follow-up interview.
The result shows that we successfully achieved all of our design goals, but it also reveals some shortcomings that can be improved in the future. 

\rev{The questionnaire includes two quantitative questions: rating the easy-to-use level and the helpful level of our system.}
When rating how our system is easy to use on a scale of $[0, 10]$ ($0$ for ``very difficult", $10$ for ``very easy"), the participants provided scores with $Md=8$, $IQR=2.25$, and more than $60\%$ of our participants' rates are $8$ or higher.
When rating the helpful level of our system on a scale of $[0, 10]$ ($0$ for ``absolutely not helpful", $10$ for ``absolutely helpful"), the participants provided scores with $Md=6$, $IQR=1.5$, and more than $75\%$ of our participants' rates are $6$ or higher. 
 
% We received an average score of 8.7 on easy-to-use and an average score of 7.2 on helpful level (fulfilled \textbf{G4});

Our observation of the user behavior and the comments we received from the interview show that most of our design goals are fulfilled well.
All participants were able to \rev{finish task T1, which means they can} generate high-level insights of models, image classes, and visual explanation methods through exploring VAC-CNN (\textbf{G1}).
One common behavior pattern of the participants was using the sortable table to investigate visual explanation results and the corresponding numerical information, through which they interprete model behaviors \rev{and answered our questions at tasks T2 and T3 (\textbf{G2, G3})}.
Most of the participants ($9$ out of $12$) mentioned that they enjoyed the smooth interface, and $4$ of them thought the real-time presents of the visual explanation results were impressive (\textbf{G4}).
``\textit{I like the way how multiple views are coordinated. I can start investigate a new model through a simple click}'', commented by participant $P4$.

However, the results also reflect some shortcomings of our system.
A few participants ($2$ out of $1$2) only had limited interactions with the distribution graph view, because they were not formiliar with clustering and felt it was hard to identify model behaviors through this visualization.
Participant $P9$ felt ``\textit{understanding a model's behavior pattern from this (view) is hard for me}''.
Some of the participants ($3$ out of $12$) mentioned in the interview that the CIH might not provide convincing results in some scenarios, and one participant thought the system can be improved \rev{by including} collective analysis towards visual explanation methods on the entire dataset.
We will discuss how to address these problems in Sec. 8.

\section{Limitations and Future Work}
\label{sec:LimitationsandFutureWork}
Through our preliminary evaluation study, we identify a few limitations of VAC-CNN.
In this section, we discuss these limitations and the corresponding future work.

\noindent\textbf{Image statistical analysis.}
The image statistical analysis functionality is supposed to support model behavior comparison when visual analytics methods fail.
However, we have found that there are many conditions when the Color Intensity Histograms can not provide convincing supplemental information for understanding model behaviors.
In the future, we plan to experiment with a new approach to image texture analysis in real-time, which should be robust and effective in various application scenarios.

\noindent\textbf{Collective model evaluation.}
Our current system includes thirteen CNN models and five visual explanation methods.
Although we support customized comparing tasks on multiple CNN models, we don't provide a collective model evaluation.
In the future, we plan to extend our work by introducing model behavior evaluation on the dataset level, with which users are able to obtain a high-level evaluation of model behaviors across the entire dataset as well as explore specific behaviors on single instances.

\noindent\textbf{Precise evaluation of qualitative comparisons.}
Our system assists researchers in combining both quantitative and qualitative analysis and allows users to update results interactively.
However, despite adding contour visualization to quantify visual explanation results, judging behavior differences of models is still largely observation-based, which could be imprecise.
In the future, we plan to incorporate quantitative measures to support evaluation, such as showing the amount of noise in the visual explanation outputs or the accuracy of the highlighted region.

\rev{
\noindent\textbf{Customization Recommendation.}
To support interpretable CNN model comparisons, our system includes multiple class-discriminative visual explanation methods and presents examples on each of them.
% Although customizable options can help users build insights from different aspects, we have found in our study that the system's efficiency would be improved if we can provide the recommendation over users' choice-making on a potential explanation methods.
% Although customizable options can support insight-building through various choices, our system could better assist users, ML novices specifically, if we can provide them with a recommendation over a potential explanation method.
Although customizable options can support insight-building by providing various tryouts, our system would be more user-friendly (ML novices in particular) if it could recommend explanation methods according to users' demands.
As future work, we plan to design recommendation strategies, such as building evaluation matrices of the visual explanation methods according to the data randomization test \cite{adebayo2018sanity}, to assist ML novices in choosing visual explanation methods.
}

\section{Conclusion}
\label{sec:Conclusion}
In this paper, we present a visual analytics system VAC-CNN (\underline{V}isual \underline{A}nalytics for \underline{C}omparing \underline{C}NNs) to assist novice ML practitioners in the comparative studies of deep Convolutional Neural Networks.
To support model interpretability, VAC-CNN integrates multiple visual explanation methods and improves the result visualization.
The system coordinates quantitative measures and informative visual explanations, and supports flexible customization of the model exploring tasks, including multi-model comparison and single-model investigation.
% We evaluate the usefulness and simplicity of VAC-CNN in supporting ML beginners through a preliminary evaluation study. 
We evaluate the usability of VAC-CNN in supporting ML beginners through a preliminary evaluation study. 
We hope our work will encourage further exploration of the inner behaviors of CNN models, and inspire the design of the next generation CNN comparison tools.
\ifCLASSOPTIONcompsoc
  \section*{Acknowledgments}
\else
  \section*{Acknowledgment}
\fi
This research is supported in part by the U.S. National Science Foundation through grant IIS-1741536 and a gift grant from Bosch Research. 
We would like to thank all the participants of our preliminary evaluation study during this challenging time.
We also want show our gratitude to Norma Gowans for narrating in our demonstration video.
We appreciated Takanori Fujiwara, Jianping (Kelvin) Li, and Qi Wu for their precious suggestions that improve this work.
We wish to extend our special thanks to anonymous reviewers for their thoughtful feedbacks and comments.


\ifCLASSOPTIONcaptionsoff
  \newpage
\fi



%\bibliographystyle{IEEEtran}


% \end{thebibliography}

\bibliographystyle{IEEEtran}
\bibliography{IEEEabrv,bibliography}


% \end{IEEEbiography}


% % insert where needed to balance the two columns on the last page with
% %\newpage

\begin{IEEEbiography}[{\includegraphics[width=1in,height=1.25in,clip,keepaspectratio]{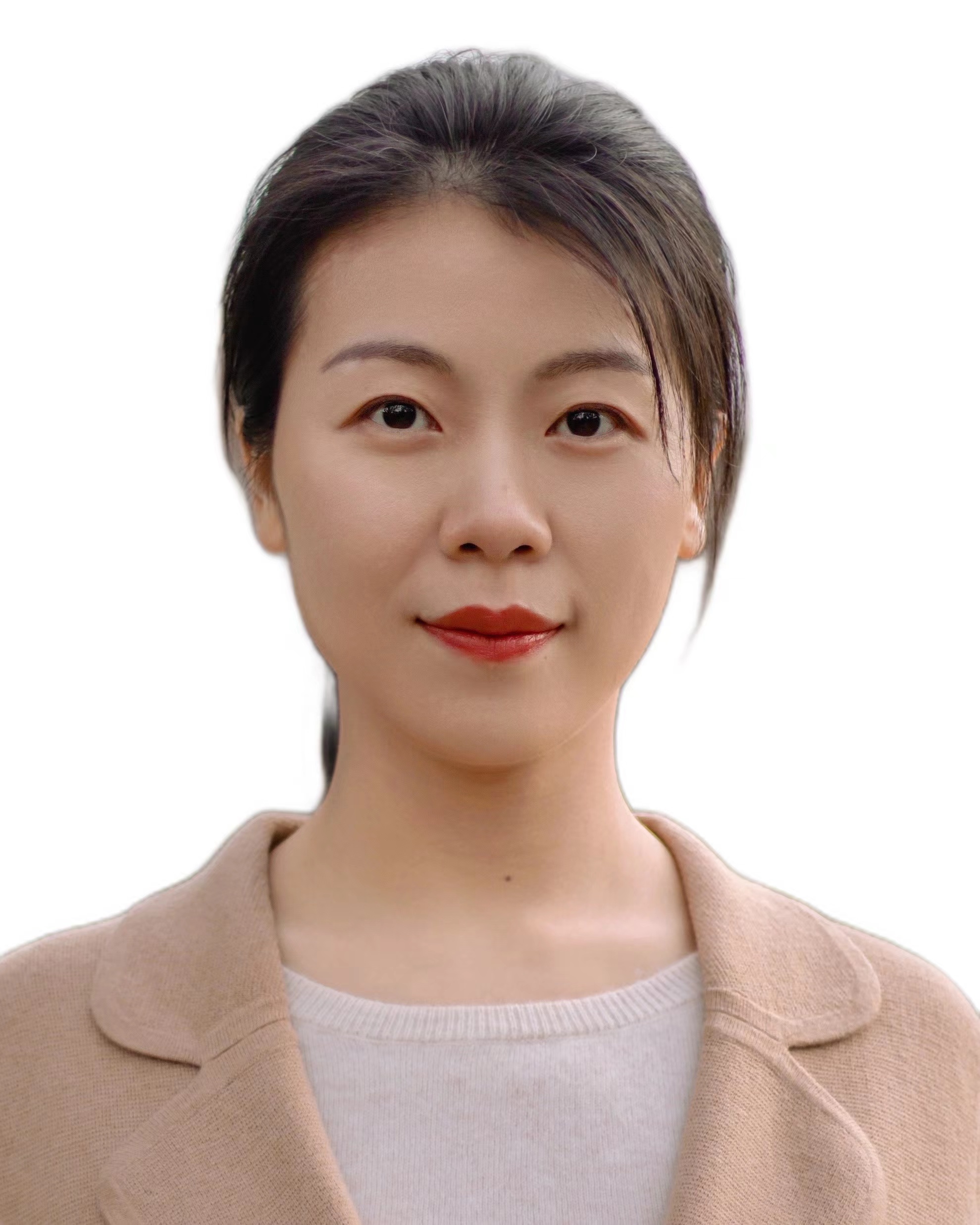}}]%
    {Xiwei Xuan} is a Ph.D. student in computer science at the University of California, Davis.
    Before UC Davis, she received the M.S. degree in electrical engineering in 2020 from the Washington University in St. Louis and received the B.S. and M.S. degrees in microelectronics from the Harbin Institute of Technology.
    Her main research interests include visual analytics, machine learning, and explainable artificial intelligence.
\end{IEEEbiography}

\begin{IEEEbiography}[{\includegraphics[width=1in,height=1.25in,clip,keepaspectratio]{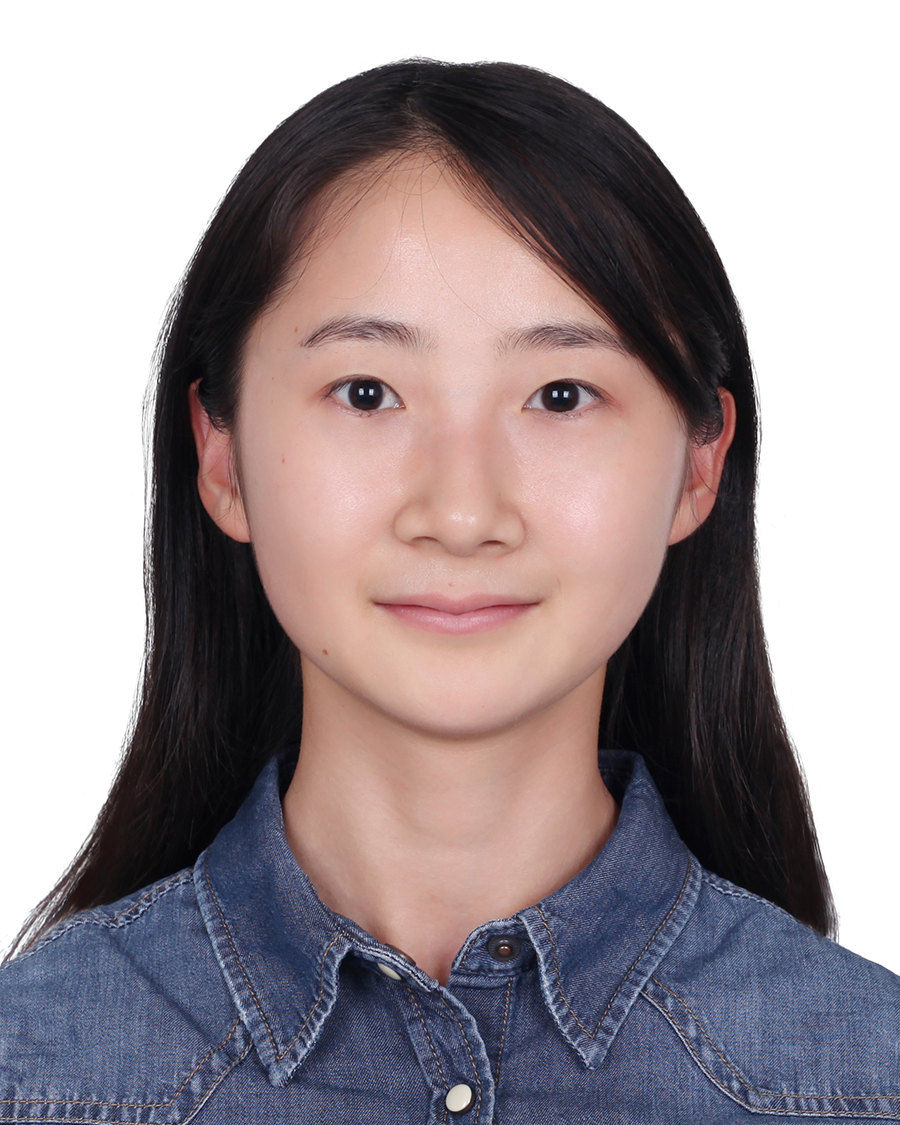}}]%
    {Xiaoyu Zhang} is a Ph.D. candidate in computer science at the University of California, Davis. She received her B.S. in digital media art from Xiamen University and her M.S. in computer science from Zhejiang University. Her research interest lies in visual analytics and information visualization. More specifically, she studies data analysis and visualization techniques to explore and exploit underlying knowledge, fact, or pattern from large text, tabular or ontological data. 
\end{IEEEbiography}

\begin{IEEEbiography}[{\includegraphics[width=1in,height=1.25in,clip,keepaspectratio]{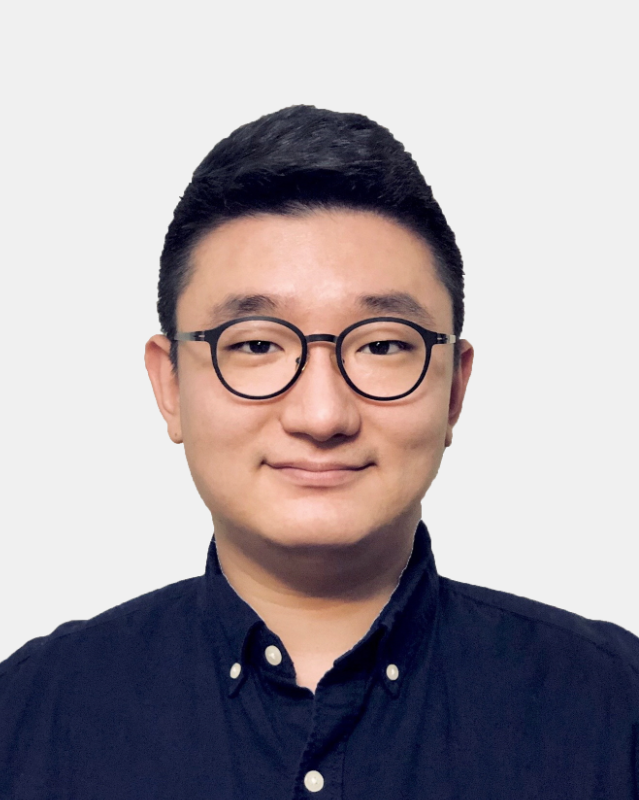}}]%
    {Oh-Hyun Kwon} received the PhD degree in computer science from the University of California, Davis in 2021. His PhD research focused on developing machine learning and immersive approaches to graph visualization. He is currently working in the industry in the areas of data visualization, visual analytics, and machine learning. 
\end{IEEEbiography}
 
\begin{IEEEbiography}[{\includegraphics[width=1in,height=1.25in,clip,keepaspectratio]{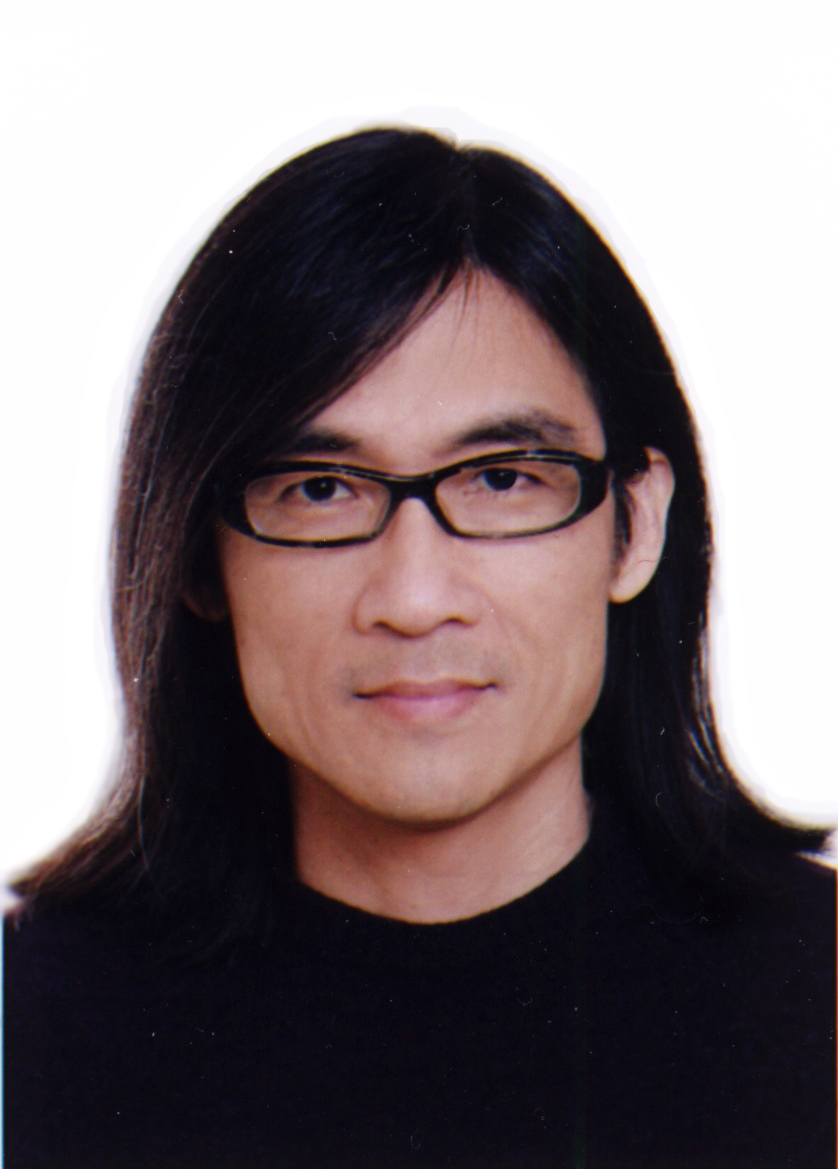}}]%
    % {Kwan-Liu Ma} is a distinguished professor of computer science at the University of California, Davis. He received his PhD degree in computer science from the University of Utah in 1993, and then worked as a staff scientist at ICASE/NASA Langley Research Center before joining UC Davis in 1999.  His research is in the intersection of data visualization, computer graphics, human-computer interaction, and high performance computing.  For his significant research accomplishments, Ma received several recognitions including elected as IEEE Fellow in 2012, recipient of the IEEE VGTC Visualization Technical Achievement Award in 2013, and inducted to IEEE Visualization Academy in 2019.
    {Kwan-Liu Ma} is a distinguished professor of computer science at the University of California, Davis. His research is in the intersection of data visualization, computer graphics, human-computer interaction, and high performance computing. For his significant research accomplishments, Ma received several recognitions including being elected as IEEE Fellow in 2012, recipient of the IEEE VGTC Visualization Technical Achievement Award in 2013, and inducted to IEEE Visualization Academy in 2019.
\end{IEEEbiography}


\vfill


\end{document}